\documentclass{article}
\usepackage{spconf,amsmath,graphicx, amssymb}

\usepackage{float}
\usepackage{gensymb} 
\usepackage[caption = false]{subfig}

\usepackage{algorithm,algorithmic,breqn} % define this before the line numbering.

\title{Crowd Flow Segmentation in Compressed Domain using CRF}
\name{Srinivas S S Kruthiventi and R. Venkatesh Babu}
\address{Video Analytics Lab\\
Supercomputer Education and Research Centre\\
Indian Institute of Science, Bangalore, India\\
{\tt\small srinivask@ssl.serc.iisc.in, venky@serc.iisc.ernet.in}%
}

\begin{document}
\ninept
\maketitle
\begin{abstract}
Crowd flow segmentation is an important step in many video surveillance tasks. In this work, we propose an algorithm for segmenting flows in H.264 compressed videos in a completely unsupervised manner. Our algorithm works on motion vectors which can be obtained by partially decoding the compressed video without extracting any additional features. Our approach is based on modelling the motion vector field as a Conditional Random Field (CRF) and obtaining oriented motion segments by finding the optimal labelling which minimises the global energy of CRF. These oriented motion segments are recursively merged based on gradient across their boundaries to obtain the final flow segments. This work in compressed domain can be easily extended to pixel domain by substituting motion vectors with motion based features like optical flow. The proposed algorithm is experimentally evaluated on a standard crowd flow dataset and its superior performance in both accuracy and computational time are demonstrated through quantitative results.
\end{abstract}
\section{Introduction}
\label{sec:intro}
Video Surveillance having become ubiquitous these days, enormous amounts of video data is captured by cameras all around us. This has made it next to impossible for any security personnel/organisation to follow and analyse these videos manually and make intelligent decisions. Fortunately, the research in computer vision is moving towards automating this process. In the past decade, automated video surveillance has become an important research topic in the field of computer vision. Research in video surveillance involves tackling problems like object/person detection, recognition, tracking, flow analysis, anomaly detection etc. 

Extracting the dominant flows present in a video forms an important preliminary step for many video surveillance tasks. Flow in a video can be defined as a dominant path along which there is significant motion throughout the video. A video can have multiple flows and neither the number of flows nor the path of each flow is known apriori. This makes the problem of flow segmentation challenging. In this work, we propose an algorithm to perform flow segmentation from videos stored in H.264 compression format \cite{h.264} in an unsupervised manner. H.264 is popular choice for video compression as it allows high resolution videos to be stored and transferred at a relatively low bandwidth. Our approach is that of segmenting the flows in the video without the need to completely decode the H.264 compressed video and without extracting any features other than motion vectors. This avoids the additional overhead of computing optical flow vectors from videos to characterise flows and makes the task of flow 
segmentation computationally minimal.

Conditional Random Fields (CRF) \cite{crf1}, which have been used extensively for vision research  in the last two decades \cite{crf2}\cite{crf3}\cite{crf4}\cite{crf5}, are known to work well for problems like image segmentation \cite{crf2}\cite{crf6}. We model the problem of flow segmentation as an optimisation problem within the framework of CRF.

The rest of the paper is organised as follows: Section 2 gives a brief overview of the recent research in flow segmentation in both compressed and pixel domains. Section 3 presents the proposed algorithm and section 4 discusses its experimental evaluation and analysis. We conclude with a summary of the proposed method in section 5.

\section{Related Work}
\label{sec:relatedWork}

In the recent past, quite a few novel approaches have been proposed for crowd analysis both in the pixel and compressed domain. In this section we discuss some of these approaches. Ali et al.~\cite{ali} proposed a Lagrangian dynamics based approach for segmentation and analysis of crowd flow. Their approach involves generating a flow field and propagating particles along them using numerical integration methods. The space-time evolution of these particles is used to setup a Finite Time Lyapunov Exponent field, which can capture the underlying Lagrangian Coherent Structure (LCS) in the flow. Dynamics and stability of the LCS reveal various flow segments present in the video.

Rodriguez et al.~\cite{rodriguez} proposed an algorithm for crowd analysis which is primarily based on prior learning of behavioural patterns from a large dataset of crowd videos. Crowd analysis is carried out by matching patches from a given test video with that of the dataset and by transferring the corresponding behavioural patterns.

Wu et al.~\cite{wu} proposed crowd motion partitioning algorithm based on representing optical flow features in salient regions as a scattered motion field. By initially making an approximation that the local crowd motion is translational in nature, the authors develop a Local-Translation Domain Segmentation (LTDS) model. They further extend this to scattered motion fields to achieve crowd motion partitioning.

The above discussed approaches work in pixel domain and involve extracting features like optical flow from the uncompressed video. In compressed domain, Gnana et al.~\cite{praveen} proposed a flow segmentation algorithm for H.264 compressed videos using motion vectors.  Their approach involves detecting region of interest in a video and clustering motion vectors extracted from those locations using Expectation Maximisation. Later the motion clusters are merged to form flows based on Bhattacharya distance between the histogram of orientation of motion vectors at the boundaries of clusters.

Again in H.264 compressed format, Biswas et al.~\cite{sovan} proposed a segmentation algorithm for crowd flow based on super-pixels. The mean motion vectors are colour coded and superpixel segmentation is performed at different scales. These segments, obtained at different scales, are merged based on boundary potential between superpixels to obtain flow segments.

\section{Proposed Method}
\label{sec:proposed}

\begin{figure}[tb]
\centering
\subfloat[]{\includegraphics[width = 1.5in]{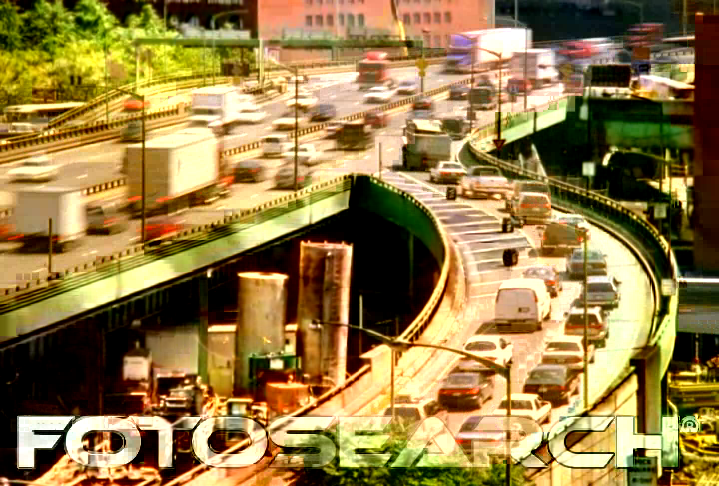}} 
\hspace{0.12in}
\subfloat[]{\includegraphics[width = 1.5in]{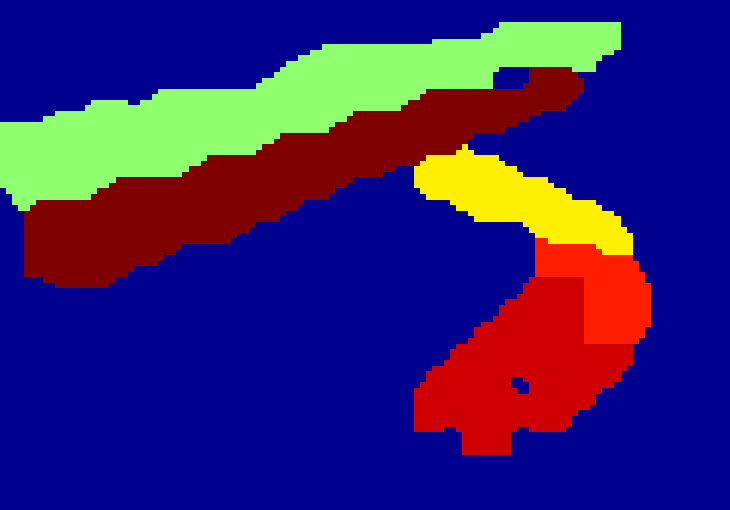}}
\hspace{0.12in}
\\
\subfloat[]{\includegraphics[width = 1.5in]{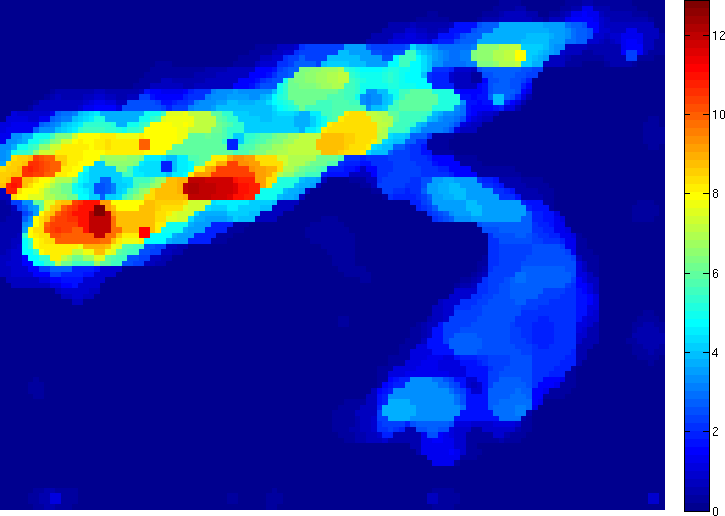}} 
\hspace{0.12in}
\subfloat[]{\includegraphics[width = 1.5in]{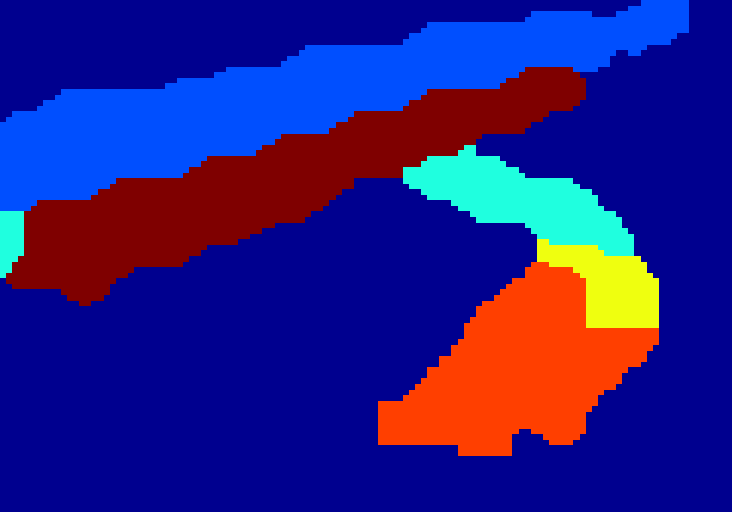}} 
\\
\subfloat[]{\includegraphics[width = 1.5in]{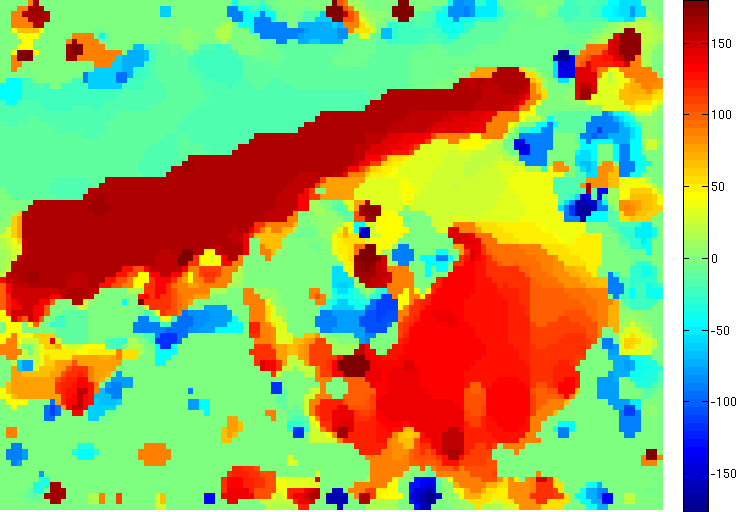}} 
\hspace{0.12in}
\subfloat[]{\includegraphics[width = 1.5in]{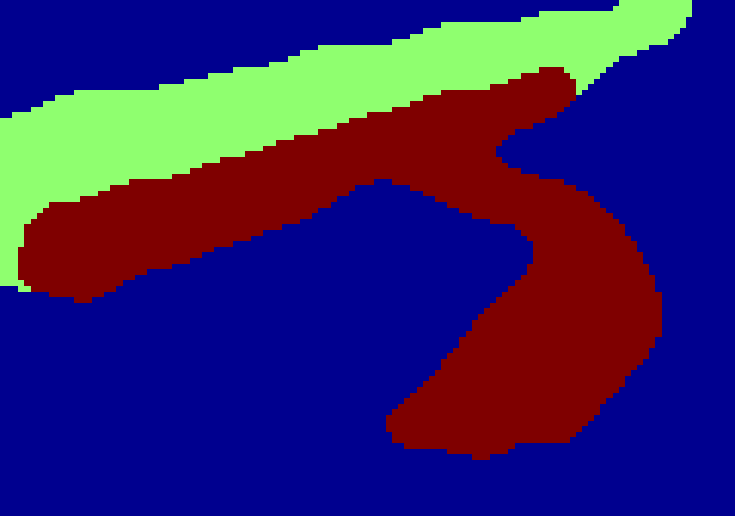}} 
\\
\caption{(a) Frame from a test sequence (c) Magnitude components of motion vector field (e) Orientation components of motion vector field (b) Segmentation result from coarse CRF (d) Segmentation result from fine CRF (f) Final flow segmentation result.}
\label{fig:demo}
\end{figure}

Our approach is based on formulating the flow segmentation problem as a CRF optimisation problem using motion vectors as features. We assign a motion vector to every 4x4 pixel block in the video by replicating motion vectors obtained from the corresponding local macro-blocks. This is to facilitate the construction of CRF on an uniform image grid. Following this, a mean motion vector field is generated by temporally averaging the motion vectors at every spatial location in the video across all frames. The magnitude and orientation components of this mean motion vector field for a test video are shown in the Fig.\ref{fig:demo} (c) and (e) respectively. The task of crowd flow segmentation in a video can be thought of as an image segmentation problem with the image being the mean motion vector field. This field can be considered as an image with two channels - magnitude and orientation of the 2D motion vectors. 

CRFs are undirected graphical models for structured prediction where the global inference is made from locally defined clique potentials. They have been rigorously used for image segmentation in the last two decades and have been proved to be great tools for this task.

\begin{figure}[htb]
\begin{minipage}{1\linewidth}
\centering
\includegraphics[scale=0.25]{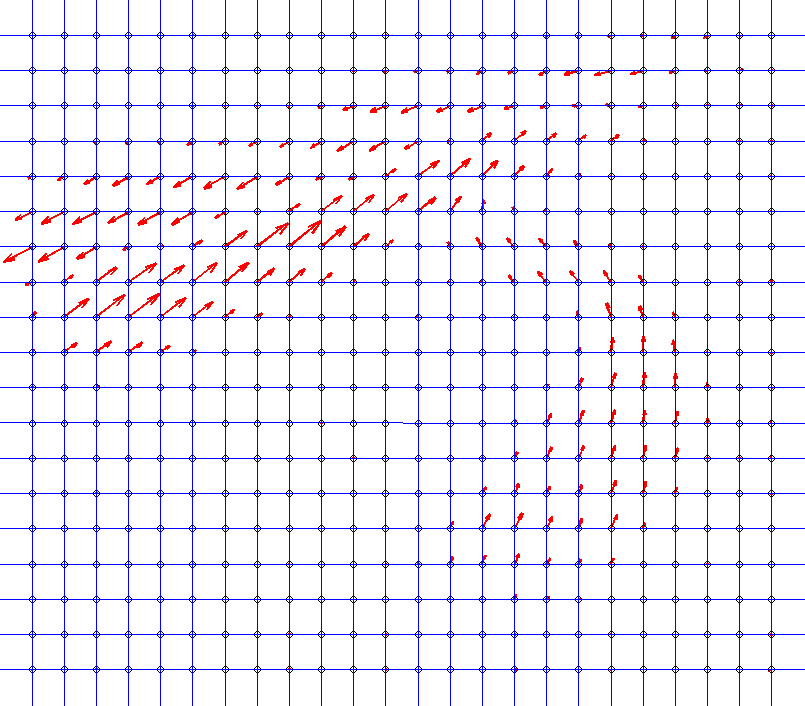}
\centerline{(a) CRF with motion feature vectors}
\end{minipage}
\begin{minipage}{0.5\linewidth}
\centering
\includegraphics[width=0.6\linewidth, height=0.6\linewidth]{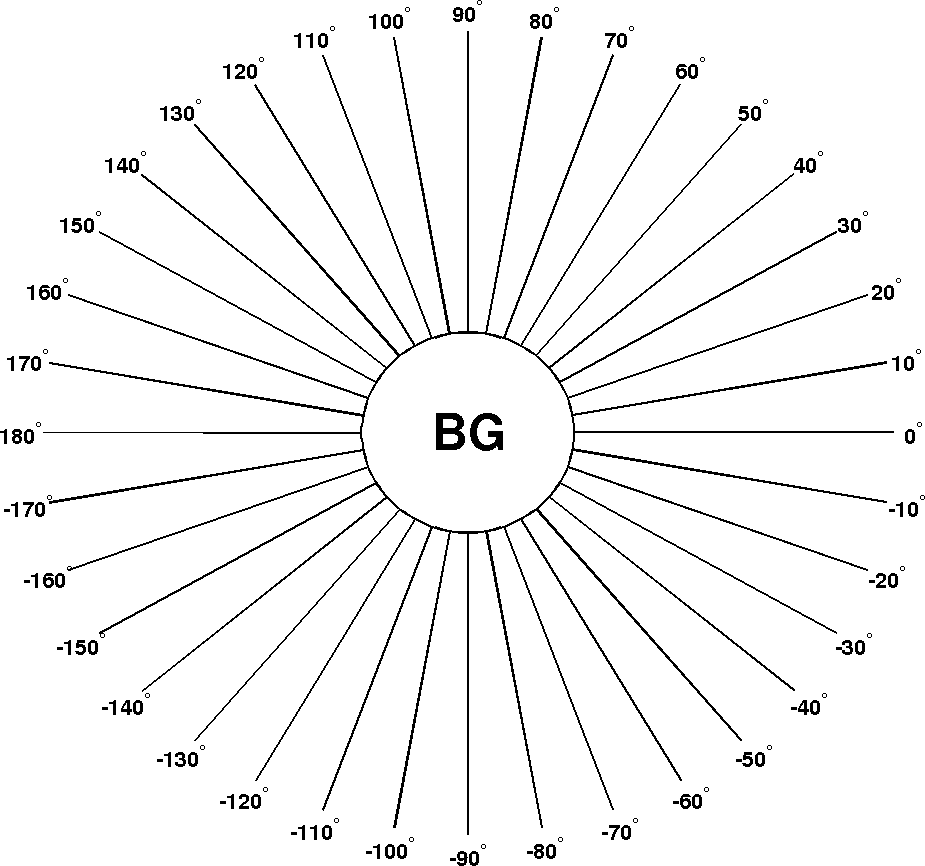}
\centerline{(b) Label orientations-coarse CRF}
\end{minipage}
\begin{minipage}{0.5\linewidth}
\centering
\includegraphics[width=0.8\linewidth, height=0.6\linewidth]{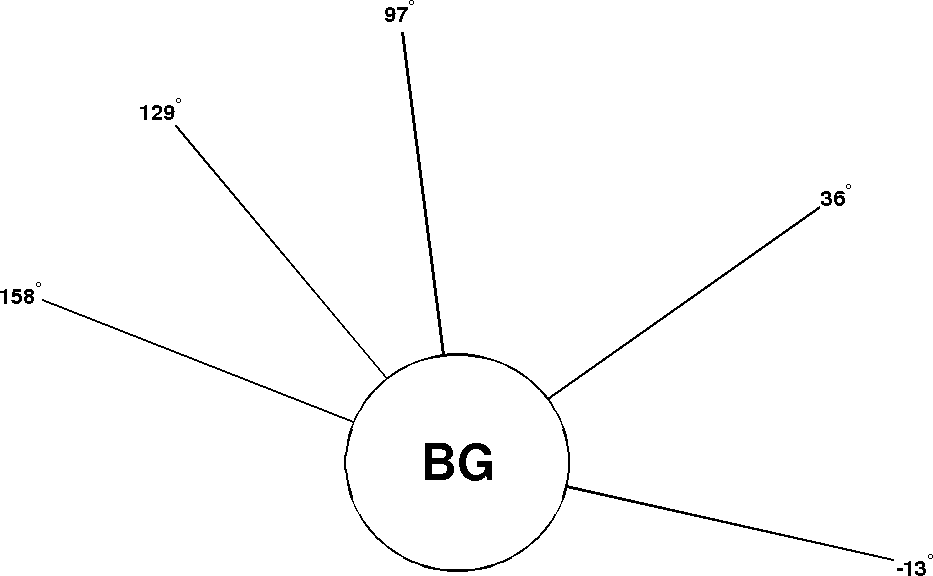}
\centerline{(c) Label orientations-fine CRF}
\end{minipage}
\caption{Formulated CRF}
\label{fig:crf}
\end{figure}

CRF is constructed on an image grid with the video's spatial dimensions and with a 4-neighbourhood connectivity. Here, each node in the CRF corresponds to the spatial location of a 4x4 pixel block in the video and is connected to its left, right, top and bottom nodes. The mean motion vector corresponding to the spatial location of each node in the CRF is taken as its feature. Let the motion vector feature corresponding to a node at location $u$ be $f^{u}$ with magnitude $f^{u}_{m}$ and orientation $f^{u}_{\theta}$. Let the label associated with this node be $x_{u}$, where $x_{u}$ is a discrete random variable. This CRF with the mean motion vector features is illustrated in Fig.\ref{fig:crf} (a).

Ideally, in this CRF formulation, each label should correspond to a flow present in the video. But the number of flows as well as their paths are unknown apriori. Hence the flow segmentation  problem is approached by initially segmenting the motion vector field based on orientation. In this, each orientation segment clusters motion vectors lying along a specific direction. Later, these motion orientation segments are merged together based on their proximity and continuity to obtain coherent flow segments. Since various motion orientations present in the video are also unknown apriori, the labels of the CRF are created to support all possible motion orientations: $-180\degree$ to $180\degree$ in steps of $10\degree$. An additional label is created to prune out the noisy motion vectors corresponding to the background in the video. This background label supports motion vectors with magnitude less than a certain threshold irrespective of their orientation. 

Specifically, for orientation based segmentation, the unary potential of a node at location $u$ with feature $f^{u}$ and label $x_{u}$ is defined as follows:

\begin{equation}
\label{eq:unary}
 \varphi_{u}(x_{u}) = \left\{ 
  \begin{array}{l l}
     0 & \;\; \text{if $x_{u} = 0 \; \& \; f^{u}_{m} < \tau$} \\
     c_1 & \;\; \text{if $x_{u} = 0 \; \& \; f^{u}_{m} \geq \tau$} \\
     c_2 & \;\; \text{if $x_{u} \neq 0 \; \& \; f^{u}_{m} < \tau$} \\
%      |f^{u}_{\theta} - \theta^{x_{u}}| & \;\; \text{if $x_{u} \neq 0 \; \& \; f^{u}_{m} \geq thresh$}
\measuredangle (f^{u}_{\theta}, \theta^{x_{u}}) & \;\; \text{if $x_{u} \neq 0 \; \& \; f^{u}_{m} \geq \tau$}
  \end{array} \right. 
\end{equation}

\begin{equation}
\label{eq:angle}
  \text{where}\;\; \measuredangle (f^{u}_{\theta}, \theta^{x_{u}}) = min(|f^{u}_{\theta} - \theta^{x_{u}}|,\;\; 360-|f^{u}_{\theta} - \theta^{x_{u}}|)
\end{equation}

Here, the label $x_{u}=0$ corresponds to the background and $\tau$ is a soft threshold on the magnitude of motion vectors to determine if they belong to the background. $c_1, c_2$ are constants determined empirically. Other labels, $x_u \neq 0$, correspond to motion along various orientations. $\theta^{x_{u}}$ is the orientation supported by the label $x_{u}$ and takes one of the values among $\{-170\degree, ..., 0\degree, ...,170\degree, 180\degree\}$. $ \measuredangle (f^{u}_{\theta}, \theta^{x_{u}})$ denotes the angle between two vectors with orientations $f^{u}_{\theta}$ , $\theta^{x_{u}}$ and is computed as given in Eq.(\ref{eq:angle}).
  
The pairwise potentials over the CRF are defined in such a way so as to ensure smooth segmentation. This is done by assigning a pairwise cost between neighbouring nodes, which take different labels, proportional to the similarity between their node features. Specifically, the pairwise potential between two neighbouring nodes $u$ and $v$ is defined as follows:

\begin{equation}
\label{eq:pairwise}
  \psi_{u,v}(x_{u},x_{v}) = \left\{ 
  \begin{array}{l l}
     0 & \;\; \text{if $x_{u} = x_{v}$} \\
     c_3*(360 - \measuredangle(f^{u}_{\theta}, f^{v}_{\theta})) & \;\; \text{if $x_{u} \neq x_{v}$}
  \end{array} \right.
\end{equation}

  \begin{algorithm}[htb] \caption{ {\bf : \textit{ Crowd Flow Segmentation}} } 

    \begin{algorithmic}

    \label{algo:flowSeg}

      \REQUIRE {Video:$V$}
      \ENSURE {Flow Segments:$\{F^0, F^1, ....., F^{N-1}\}$}
      \STATE
      
      \STATE \% Extract mean motion vector field from V
      \STATE $MV = MeanMotionVectors(V)$
      \STATE
      
      \STATE Labels: $0, 1, ..., K-1$
      \STATE \%Label 0 corresponds to background $\&$ supports motion vectors of magnitude less than a threshold
      \STATE
      
      \STATE \% $\theta_{coarse}^i$ : Orientation supported by label i
      \STATE $\theta_{coarse}^i = -180 + i*10 \quad \forall i\in[1\;\;K-1]$
      \STATE
	
      \STATE \%Extract Coarse Orientation Segments 
      \STATE \%Unary and Pairwise costs are defined in Eq.(\ref{eq:unary}) and Eq.(\ref{eq:pairwise})
      \STATE $\{S_{coarse}^0,... S_{coarse}^{L-1}\} = CRFoptimisation(MV,\boldsymbol\theta_{\bf{coarse}})$
      \STATE
      
      \STATE $i = 1$
      \FOR {$l=0~\to~L-1$}
      \IF {$(|S_{coarse}^l| > size_{thresh})$}
      \STATE $\theta_{fine}^i = MeanOrientation(S_{coarse}^l)$
      \STATE $i=i+1$
      \ENDIF
      \ENDFOR
      \STATE
      
	\STATE \%Extract Fine Orientation Segments 
	\STATE $\{S_{fine}^0,... S_{fine}^{M-1}\} = CRFoptimisation(MV, \boldsymbol\theta_{\bf{fine}})$
	\STATE
	
	\STATE \%Extract Flow Segments 
	\STATE $\{F^0, F^1,... F^{N-1}\} = Merge(S_{fine}^0,... S_{fine}^{M-1})$
	\STATE
	
    \end{algorithmic} 
  \end{algorithm}

With the unary and pairwise potentials as defined in Eq.(\ref{eq:unary}) and Eq.(\ref{eq:pairwise}), the total energy of the CRF is the sum of unary and pairwise terms:\\

\begin{equation}
\label{eq:energy}
E(\boldsymbol x) =  \sum_{\substack{u}} \varphi_{u}(x_{u}) + \sum_{\substack{u,v \\ u \neq v}} \psi_{u,v}(x_{u}, x_{v})
\end{equation}

Solving for the CRF, thus formulated, is equivalent to finding a labelling $\boldsymbol x^* = [..., x_{u}, ..., x_{v}, ...]$, which minimises the global energy  $E(\boldsymbol x)$ defined in Eq.(\ref{eq:energy}). The optimal labelling assigns a label to each node in the image grid, thus assigning it into either a background segment or a segment with a specific orientation. The oriented motion segmentation result obtained is shown in Fig.\ref{fig:demo} (b).

Finding the exact solution for the minimum energy labelling problem is NP hard. In this work, an approximate solution for the CRF labelling is found out using the graph cuts based algorithm proposed in the works of \cite{gco1, gco2, gco3, gco4}. Their algorithm converges quickly for grid graphs to a local minima by allowing large moves whenever possible.  

The motion segmentation, so obtained, is coarse and may not be very accurate. This is because the orientations supported by the CRF labels({$-170\degree, -160\degree, ..., 180\degree$}), need not closely align with the actual orientations present in the motion vector field. In order to further refine this segmentation, we formulate a fine CRF. The labels for this fine-CRF are obtained by taking the mean orientation of motion vectors contained in each coarse segment. Here we consider only segments whose size is greater than a certain threshold. This helps in eliminating noisy segments. This fine CRF is solved with the same unary and pairwise potentials as in Eq.(\ref{eq:unary}) and Eq.(\ref{eq:pairwise}) with $\theta^{x_{u}}$ corresponding to the newly calculated orientations. The label orientations corresponding to the coarse CRF and the fine CRF are shown in Fig.\ref{fig:crf} (b) and (c) respectively. The refined motion segmentation obtained after solving this fine CRF is shown in Fig.\ref{fig:demo} (d).

The final flow segmentation is obtained by appropriately merging the refined oriented motion segments. For this purpose, we create a gradient image of the orientation channel of the motion vector field. Now, we consider the mean gradient along the boundary joining the two segments which are considered for merging. If this mean gradient is less than a certain threshold, the two segments are merged. The entire algorithm is summarised in Algorithm. \ref{algo:flowSeg}. The final flow segments obtained are shown in Fig.\ref{fig:demo} (f).

\section{Experiments}
\label{sec:experiments}

The proposed method is evaluated on the flow dataset provided by Ali et al.~\cite{ali}. The videos of this dataset have dense flows in both traffic and crowd scenarios. Since these videos are not originally present in H.264 format, we have followed the same procedure as Biswas et al.~\cite{sovan} for encoding. Specifically, the video is encoded into H.264 baseline with only I \& P frames. One reference frame is considered with the Group of Pictures length set to 30. As mentioned in \cite{sovan}, this baseline profile is ideal for extracting motion vectors on-the-fly with low latency. The motion vectors extracted from the encoded video can come from varying macro-block sizes (from 4x4 to 16x16). The motion vectors obtained from bigger macro-blocks are replicated to their constituent 4x4 blocks to maintain grid uniformity and facilitate comparison of results with \cite{sovan}. 

\begin{figure*}[t]
\centering
Test Sequences	\hspace{0.8in} Ground Truth \hspace{0.8in}  Biswas et al.\cite{sovan}	\hspace{1in}  Proposed
\\
\subfloat{\includegraphics[width = 1.5in, height=1in]{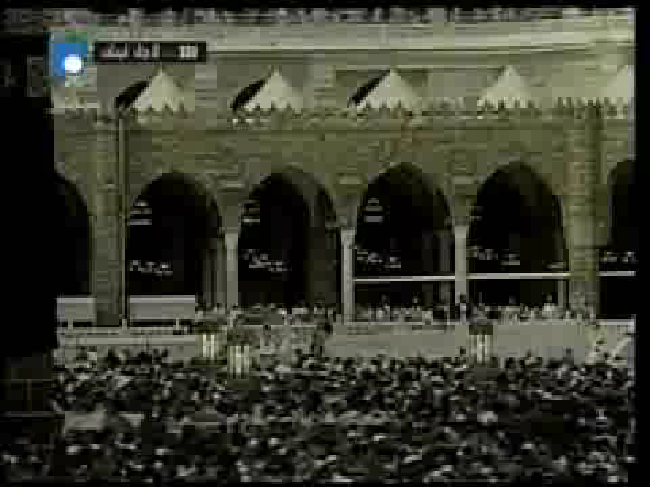}} 
\hspace{0.1in}
\subfloat{\includegraphics[width = 1.5in, height=1in]{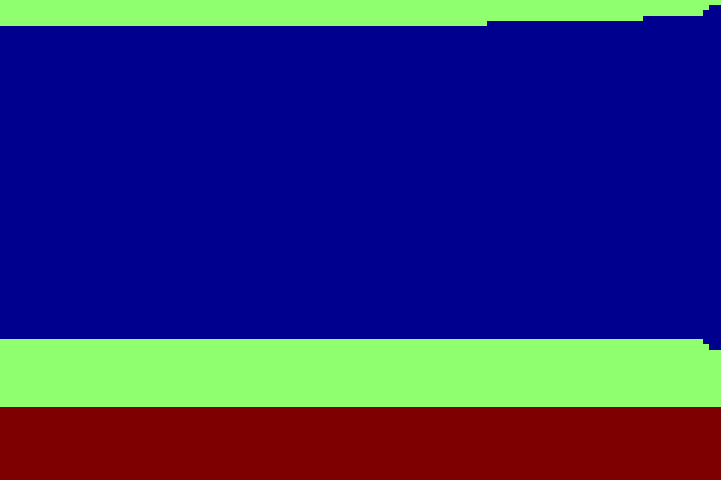}}
\hspace{0.1in}
\subfloat{\includegraphics[width = 1.5in, height=1in]{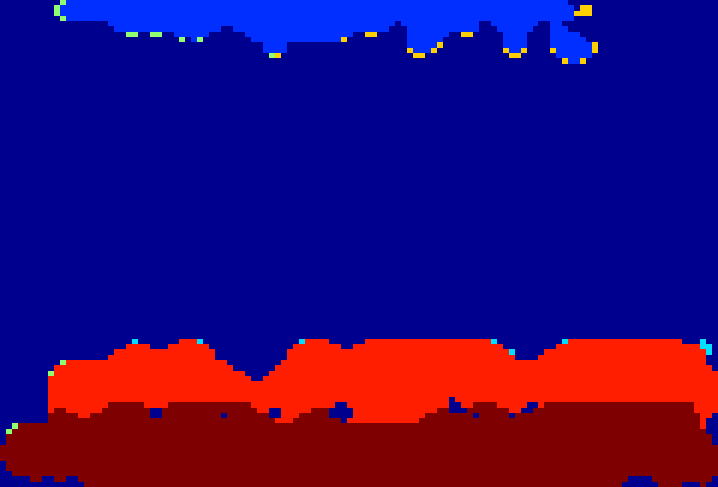}}
\hspace{0.1in}
\subfloat{\includegraphics[width = 1.5in, height=1in]{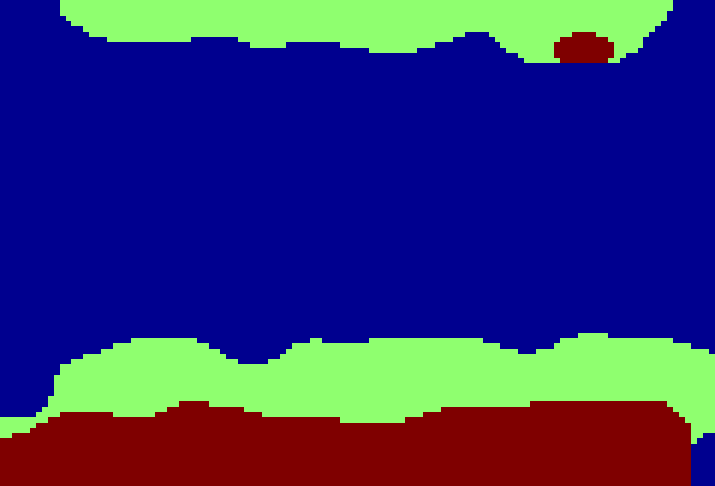}}
\\
(a) Sequence 3
\vspace{-0.3cm}
\\
\subfloat{\includegraphics[width = 1.5in, height=1in]{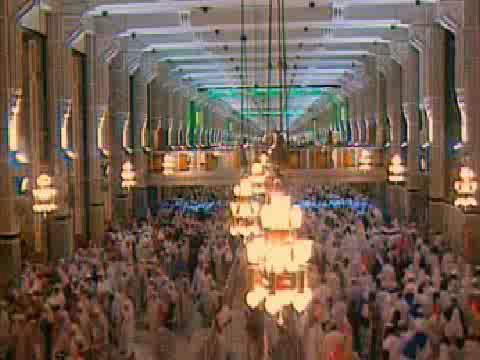}} 
\hspace{0.1in}
\subfloat{\includegraphics[width = 1.5in, height=1in]{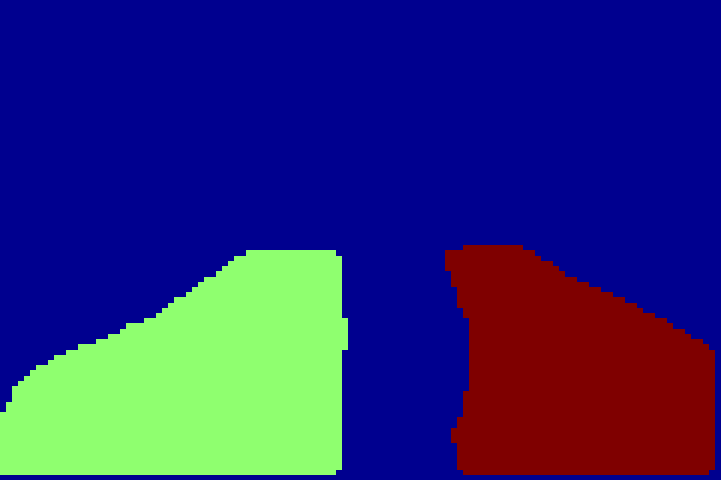}}
\hspace{0.1in}
\subfloat{\includegraphics[width = 1.5in, height=1in]{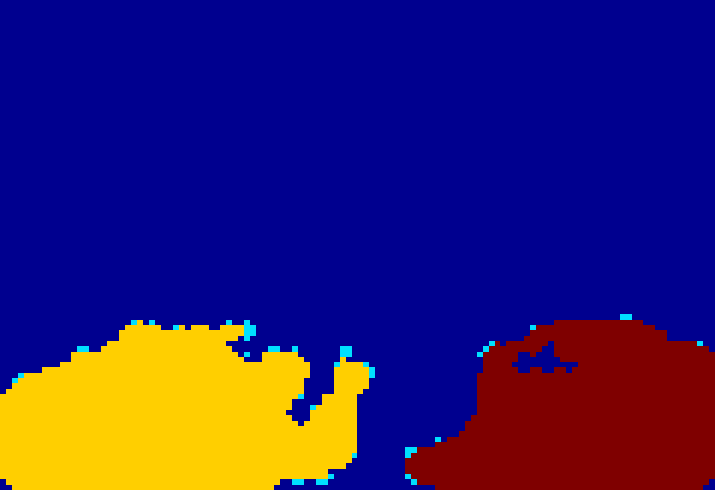}}
\hspace{0.1in}
\subfloat{\includegraphics[width = 1.5in, height=1in]{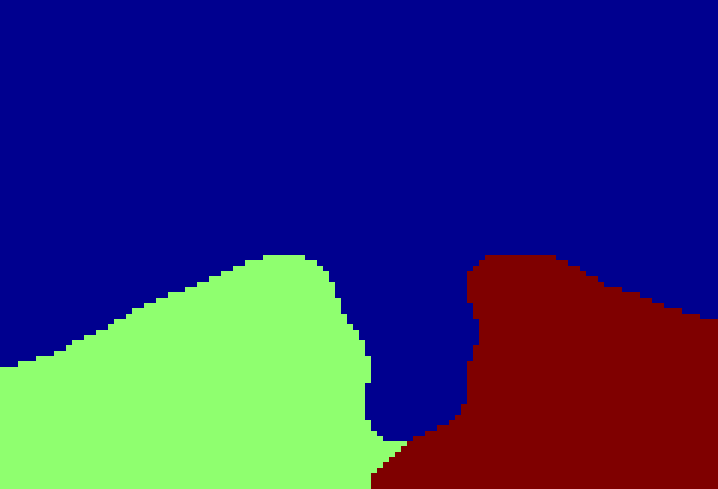}}
\\
(b) Sequence 6
\vspace{-0.3cm}
\\
\subfloat{\includegraphics[width = 1.5in, height=1in]{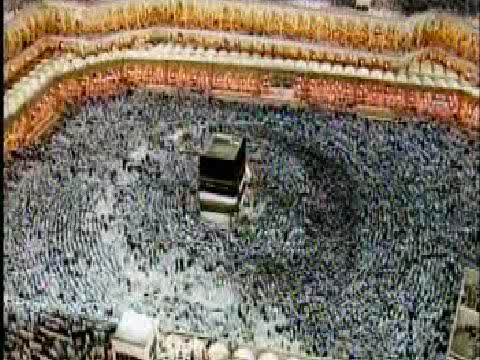}} 
\hspace{0.1in}
\subfloat{\includegraphics[width = 1.5in, height=1in]{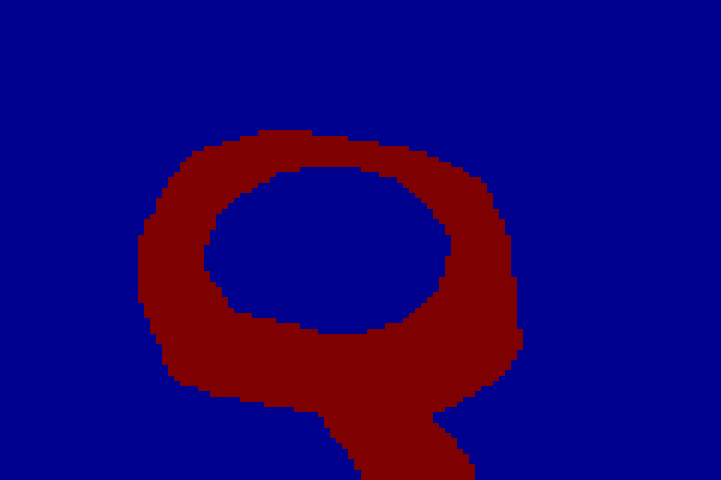}}
\hspace{0.1in}
\subfloat{\includegraphics[width = 1.5in, height=1in]{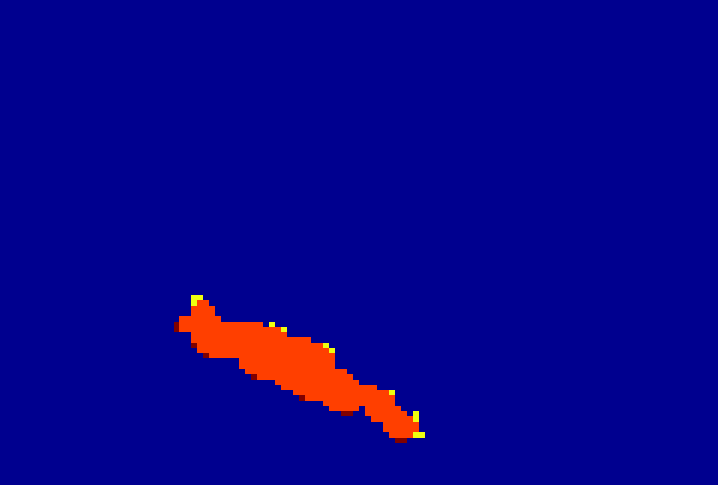}}
\hspace{0.1in}
\subfloat{\includegraphics[width = 1.5in, height=1in]{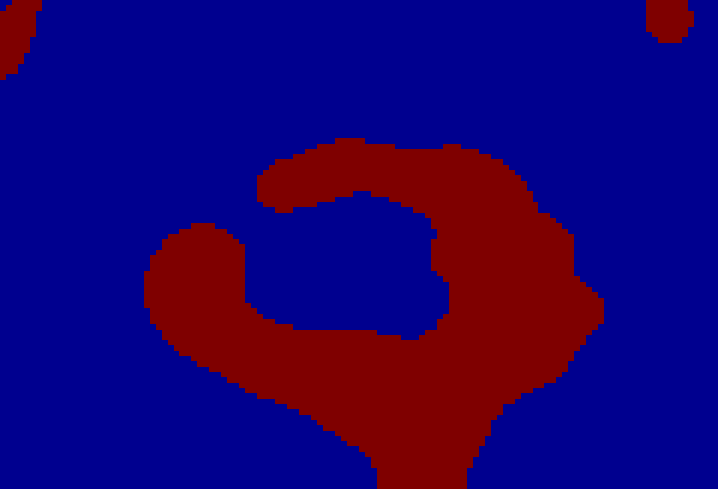}}
\\
(c) Sequence 7
\caption{Qualitative results for crowd flow segmentation. (More results at http://val.serc.iisc.ernet.in/srinivas/CRFFlowSeg.html)}
\label{fig:results}
\end{figure*}

The flow segments obtained using the proposed algorithm are quantitatively evaluated by comparing against the ground-truth segments and using the Jaccard similarity measure. Let the ground-truth segmentation be $A$ and the output of the proposed algorithm be $B$. The Jaccard measure, which is the value of intersection over union, for $A$ and $B$ can be computed as 

\begin{equation}
 J(A, B) = \frac{|A \cap B|}{|A \cup B|}
\end{equation}

Here the intersection represents the number of non-zero labelled pixel locations which match in labelling $A$ and labelling $B$. The union represents the number of pixel locations which are assigned a non-zero label in either $A$ or $B$ or both.

\begin{table}[]
\caption{Jaccard Similarity Measure with Ground Truth}
\begin{center}
\label{tab:accuracy}
\begin{tabular}{|l|c|c|c|r|}
\hline
Test Sequences & Ali et al.\cite{ali} & Biswas et al.\cite{sovan} & Proposed\\
\hline\hline
Sequence 1  	& {0.63} & {0.60} & {\bf0.90}\\
\hline
Sequence 2  	& {0.28} & {\bf0.67} & {0.66}\\
\hline
Sequence 3  	& {0.57} & {0.74} & {\bf0.75}\\
\hline
Sequence 4  	& {0.67} & {0.68} & {\bf0.68}\\
\hline
Sequence 5  	& {\bf0.78} & {0.24} & {0.46}\\
\hline
Sequence 6  	& {0.41} & {0.62} & {\bf0.81}\\
\hline
Sequence 7  	& {\bf0.60} & {0.15} & {0.53}\\
\hline
\end{tabular}
\end{center}
\end{table}

The quantitative and qualitative results are shown in Table. \ref{tab:accuracy} and Fig.\ref{fig:results} respectively. The timing results presented in Table. \ref{tab:speed} are based on experiments performed in MATLAB on a 3.4 GHz 64-bit Linux system with 24GB RAM.

In Sequence 5, the frame size is 188$\times$144 compared to 480$\times$360 for the other videos. Here the motion vectors could not capture motion accurately enough resulting in bad performance. As long as the motion is well captured, the proposed approach is shown to perform better or equivalent to \cite{ali}, a pixel domain based approach. Computationally, \cite{ali} takes around 30 sec for each sequence which is two orders of magnitude slower compared to the proposed method.

\begin{table}
\caption{Computational Time (in sec)}
\begin{center}
\label{tab:speed}
\begin{tabular}{|l|c|c|c|r|}
\hline
Video Sequences & Biswas et al.\cite{sovan} & Proposed\\
\hline\hline
Sequence 1  	& {4.96} & {\bf0.20}\\
\hline
Sequence 2  	& {5.08} & {\bf0.31}\\
\hline
Sequence 3  	& {4.66} & {\bf0.23}\\
\hline
Sequence 4  	& {4.49} & {\bf0.33}\\
\hline
Sequence 5  	& {4.32} & {\bf0.08}\\
\hline
Sequence 6  	& {5.32} & {\bf0.31}\\
\hline
Sequence 7  	& {4.95} & {\bf0.38}\\
\hline
\end{tabular}
\end{center}
\end{table}

\section{Conclusion}
\label{sec:conclusion}

In this work, we have proposed an algorithm for crowd flow segmentation in the framework of CRFs. The node features for CRF are taken to be the motion vectors and unary and pairwise terms are so defined to obtain cluster segments corresponding to motion along various orientations. Initially, we consider the labels for CRF to support all possible orientations in the $360^{0}$ plane and later refine them based on orientations present in the video. The refined orientation segments are recursively merged to obtain the final flow segments. Our method can also be applied in pixel domain by just replacing the motion vectors with optical flow vectors.

One drawback of the proposed approach and other recent methods \cite{praveen, sovan} is their inability to handle intersecting flows. This work can be extended to segment time-varying flows by constructing a multi-modal model at every spatial location as opposed to just the mean statistics.

\section{Acknowledgement}
This work was supported by Defence Research Development Laboratory (DRDO), project No. DRDO0672.

\bibliographystyle{IEEEbib}
\bibliography{strings,refs}

\end{document}